\documentclass[10pt]{article}
\usepackage[preprint]{neurips_2024}

\usepackage[utf8]{inputenc} 
\usepackage[T1]{fontenc}    
\usepackage{hyperref}       
\usepackage{url}            
\usepackage{booktabs}       
\usepackage{amsfonts}       
\usepackage{nicefrac}       
\usepackage{microtype}      
\usepackage{xcolor}         
\usepackage{graphicx}
\usepackage[numbers]{natbib}

\title{Formatting Instructions For NeurIPS 2024}

\begin{document}

\title{Automated Image Captioning with CNNs and Transformers}
\author{%
  Joshua A.~Cahyono\thanks{These authors contributed equally to this work.} \\
  College of Computing and Data Science\\
  Nanyang Technological University\\
  \texttt{jo0001no@e.ntu.edu.sg}\\
  \And 
  Jeremy N.~Jusuf\footnotemark[1] \\
  College of Computing and Data Science\\
  Nanyang Technological University\\
  \texttt{jere0048@e.ntu.edu.sg}\\
}

\maketitle
\footnotetext[1]{These authors contributed equally to this work.}
\footnotetext[2]{Source code is available at \href{https://github.com/JeremyNathanJusuf/image-captioning}{GitHub}.}

\begin{abstract}
  This project aims to create an automated image captioning system that generates natural language descriptions for input images by integrating techniques from computer vision and natural language processing. We employ various different techniques, ranging from CNN-RNN to the more advanced transformer-based techniques. Training is carried out on image datasets paired with descriptive captions, and model performance will be evaluated using established metrics such as BLEU, METEOR, and CIDEr. The project will also involve experimentation with advanced attention mechanisms, comparisons of different architectural choices, and hyperparameter optimization to refine captioning accuracy and overall system effectiveness.
\end{abstract}

\section{Introduction}
The task of captioning and image lies at the intersection of computer vision and natural language processing, two key domains in artificial intelligence. The goal of image captioning systems is to produce accurate, meaningful, and contextually appropriate descriptions of the contents of an image. This capability has significant applications, including helping visually impaired people describe their surroundings, enhancing search engine capabilities for images, and improving content organization in social media and digital asset management.

Earlier approaches to image captioning relied on rule-based or retrieval-based methods that required substantial manual effort to create templates and caption databases. However, recent advances in deep learning have enabled the development of automated captioning systems that can generate more flexible and varied descriptions by learning directly from image-caption pairs. Modern approaches typically use a mixture of neural networks, such as Convolutional Neural Networks (CNNs) for extracting visual features from images and Recurrent Neural Networks (RNNs) or Transformer architectures to model and generate descriptive sentences in natural language.

This project focuses on building an image captioning system using a series of progressively advanced models: \textbf{CNN-RNN}, CNN-Attention \textbf{(CNN-Attn)}, YOLO with CNN and Attention \textbf{(YOLOCNN-Attn)}, Vision Transformer with Attention \textbf{(ViT-Attn)}, and finally a \textbf{novel hybrid} Vision Transformer-CNN with Attention \textbf{(ViTCNN-Attn)}. 

The system will be trained on 2 datasets (\textbf{MSCOCO} and \textbf{Flickr-30k}) containing images and their corresponding captions, enabling it to learn from diverse visual contexts and linguistic expressions. The performance of the model will be assessed using standard metrics such as \textbf{BLEU}, \textbf{METEOR}, and \textbf{CIDEr}, which evaluate the accuracy and fluency of the generated captions. Additionally, we will explore various enhancements, including experimenting with different attention mechanisms, optimizing hyperparameters, and comparing alternative architectural configurations, with the objective of producing high-quality captions that closely align with human expectations.

By the end of this project, we aim to establish a comprehensive understanding of image captioning models, from basic CNN-RNN approaches to advanced Transformer-based systems, and assess their effectiveness in generating natural and contextually accurate descriptions for images.

\section{Related Work}
\subsection{CNN-RNN Architecture for Image Captioning}

\begin{figure}[b]
  \centering
  \includegraphics[width=0.5\textwidth]{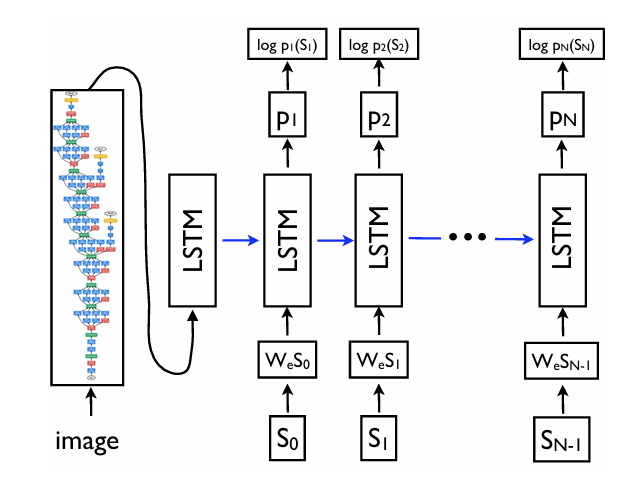}
  \caption{CNN-RNN Architecture for Image Captioning}
   \label{fig:cnn-rnn}
\end{figure}

The CNN-RNN architecture is a pioneering approach for generating image captions by combining Convolutional Neural Networks (CNNs) for visual representation and Recurrent Neural Networks (RNNs) for sequential text generation.

In this architecture, a CNN serves as the "encoder" to extract rich features from the input image, which are subsequently used by an RNN to generate a natural language description. The CNN, often pre-trained on large image classification datasets (e.g., VGG16, Inception), converts an image into a fixed-length feature vector. These features are then fed into an RNN decoder, typically employing Long Short-Term Memory (LSTM) units, which are capable of handling long-term dependencies crucial for language generation. The LSTM iteratively generates words one by one until a complete caption is formed. (see Figure \ref{fig:cnn-rnn})

The core idea of the CNN-RNN model for image captioning was demonstrated by Vinyals et al. \cite{vinyals2015show} with their "Show and Tell" model. Unlike traditional rule-based methods that required engineered features and rigid language rules, this model takes advantage of deep learning's ability to learn both visual and linguistic representations simultaneously, resulting in more accurate and contextually relevant descriptions. 

Overall, this CNN-RNN approach set a new benchmark in the field of image captioning by providing a unified, end-to-end trainable architecture that greatly outperformed traditional captioning systems. The combination of CNNs for visual encoding and LSTMs for language decoding effectively bridged the gap between vision and natural language, showcasing deep learning's power in multimodal learning tasks.

\subsection{YOLO-CNN-RNN}
The YOLO-CNN-RNN model builds on the strengths of the CNN-RNN architecture by incorporating the YOLO (You Only Look Once) object detection framework. YOLO, introduced by Redmon et al. \cite{redmon2016you} allows for efficient detection of key objects in an image, enhancing the model's ability to focus on important regions during caption generation. By combining YOLO with CNN and RNN, this approach aims to provide more accurate and informative captions, especially for images containing multiple distinct objects. The YOLO component is used to detect bounding boxes and classify the objects in real time, allowing the subsequent RNN to generate captions that specifically refer to these detected objects. This integration aims to improve the model's interpretability and its ability to provide detailed captions in cluttered scenes where multiple elements are present.

\subsection{CNN-Attn}

In traditional RNN-based architectures, the vanishing gradient problem is a major challenge, especially when dealing with long sequences or high-dimensional image data. To address this issue, we incorporate \textbf{Attention Mechanism}. The introduction of an attention mechanism is crucial for alleviating the vanishing gradient problem. By dynamically focusing on specific image regions at each step, the model avoids needing to remember all image information throughout the sequence generation. This reduces the burden on LSTMs for maintaining gradients across long temporal dependencies, thereby mitigating the vanishing gradient problem.

This approach was significantly advanced by Xu et al. \cite{xu2015show} with their "Show, Attend and Tell" model, which introduced an attention-based mechanism to the image captioning task, allowing the model to focus on different parts of the image when generating different words.

\subsection{ViT-GPT}
The ViT-GPT model leverages Vision Transformers (ViT) for visual feature extraction and integrates it with a GPT-like architecture for generating captions. Vision Transformers, as introduced by Dosovitskiy et al. \cite{dosovitskiy2020image}, are known for their ability to capture global dependencies in images through self-attention, making them highly effective in understanding complex visual patterns. By combining ViT with a GPT-style language model, this approach aims to generate high-quality captions that are both contextually rich and coherent. The ViT component divides the image into patches, and each patch is treated as a token, allowing the model to learn relationships between different parts of the image effectively. The integration with GPT, inspired by Radford et al. \cite{radford2018improving}, provides a powerful language model capable of generating fluent and context-aware captions. The ViT-GPT model is particularly advantageous for handling diverse and intricate visual content, as it leverages the power of Transformers to process both visual and textual information comprehensively, enabling the generation of captions that reflect a deeper understanding of image content.

\section{Methodology}
\subsection{Datasets}

\begin{table}
  \caption{Statistics of the Datasets}
  \label{table-1}
  \centering
  \begin{tabular}{cccccc}
    \toprule
    Name         & Train Split & Val Split & Test Split & Total Images & Vocab Size \\
    \midrule
    \textbf{MS COCO}      & 82,783         & 36,453          & 4,051     & 164,062 & 8843     \\
    \textbf{Flickr30k}    & 27,173         & 3,020            & 1,590      & 31,783 & 7229    \\
    \bottomrule
  \end{tabular}
\end{table}

We evaluated our approach using two commonly utilized datasets for image captioning: MS COCO \cite{lin2014microsoft} and Flickr30k \cite{young2014image}. Table \ref{table-1} presents concise statistics of these datasets. Both datasets were gathered from the Flickr photo-sharing platform and comprise real-world images, each annotated with five descriptions provided by human annotators. For both datasets, we use an 85-10-5 split ratio for training, validation, and test respectively. We only take the vocabulary from the training set and set anything not in the vocabulary list as unknown.

\subsection{Objective and Loss Function}

We want to maximize the probability of the correct description given the image by using the following formulation:

\begin{equation}
    \theta = \arg\max_\theta \log p(S | I; \theta)
    \label{eq:maximize_log_prob}
\end{equation}

where \( \theta \) represents the parameters of our model, \( I \) is an image, and \( S \) is its correct caption. Since \( S \) represents any sentence, its length is unbounded. Thus, it is common to apply the chain rule to model the joint probability over \( S_0, \ldots, S_N \), where \( N \) is the length of this particular example, as:

\begin{equation}
    \log p(S | I) = \sum_{t=0}^{N} \log p(S_t | S_0, \ldots, S_{t-1}, I)
    \label{eq:chain_rule}
\end{equation}

At training time, \( (S, I) \) is a training example pair, and we optimize the sum of the log probabilities as described in Equation \ref{eq:chain_rule} over the whole training set using stochastic gradient descent (SGD).

To optimize this objective, we use the \textbf{cross-entropy loss}, which is well-suited for classification tasks, such as predicting the next word in a sequence from a fixed vocabulary. Cross-entropy loss measures the difference between the predicted probability distribution over the vocabulary and the actual distribution, typically represented as a one-hot encoded vector. Specifically, for each time step \( t \), we calculate:

\begin{equation}
    L_t = - \sum_{w \in V} y_{t,w} \log(\hat{y}_{t,w})
    \label{eq:cross_entropy}
\end{equation}

where:

\begin{itemize}
    \item \( V \) represents the vocabulary.
    \item \( y_{t,w} \) is the actual one-hot encoded vector for the correct word at time step \( t \).
    \item \( \hat{y}_{t,w} \) is the predicted probability for each word \( w \) in the vocabulary at time step \( t \).
\end{itemize}

The cross-entropy loss \( L_t \) penalizes the model if the predicted probability diverges from the true label, encouraging the model to assign higher probabilities to the correct words during sequence generation. By summing over all time steps \( t \), we minimize the total loss across the sequence, thereby improving the model’s ability to predict words from a given vocabulary accurately.

\subsection{Evaluation Metrics}

In our image captioning experiment, we utilized cross-entropy loss, as each token in the caption generation task is framed as a classification problem. For evaluation metrics, we employed BLEU \cite{papineni2002bleu}, METEOR \cite{denkowski2014meteor}, and CIDEr \cite{vedantam2015cider}.

\begin{itemize}
    \item \textbf{BLEU} (Bilingual Evaluation Understudy) measures the geometric mean overlap of n-grams between the generated and reference captions, making it effective for assessing precision. BLEU has been shown to correlate well with human evaluations when comparing the generated captions against human-annotated references. In particular, we use \textbf{BLEU-4}, which considers the geometric mean of 1-gram up to 4-gram overlaps to better capture the quality of longer phrases in the generated captions.
    \item \textbf{METEOR} (Metric for Evaluation of Translation with Explicit ORdering) goes beyond simple n-gram matching by considering synonyms, stemming, and word order, thus providing a more nuanced evaluation. By accounting for linguistic diversity and meaning, METEOR offers better alignment with human judgments in some contexts, especially when variations in word choice are allowed.
    \item \textbf{CIDEr} (Consensus-based Image Description Evaluation) emphasizes consensus among multiple human-annotated references, making it particularly useful for capturing the relevance and informativeness of the generated captions. CIDEr computes a weighted average of term frequency-inverse document frequency (TF-IDF) values across reference captions, ensuring that generated captions are not only grammatically correct but also semantically relevant based on common human descriptions.
\end{itemize}

To ensure that the evaluation is fair and consistent across different metrics, we use the \textbf{NLGMetricVerse} library, which provides standardized and reliable implementations of these evaluation metrics, thereby minimizing discrepancies and maintaining comparability in the results.

\subsection{Training Details}

In this work, we present a robust and modular pipeline for the efficient training and evaluation of image captioning models. 
The process begins with a data loading module that ingests the dataset and partitions it into training, validation, 
and test subsets. During preprocessing, images are paired with their corresponding reference captions to form distinct 
training samples, leveraging \textbf{PyTorch’s Dataset and DataLoader classes} for efficient data handling. We also ensure 
that appropriate image transformations are applied, tailored to the specific requirements of the chosen image encoder, 
thus optimizing the input for pretrained feature extractors. This comprehensive approach allows for streamlined data 
processing and effective utilization of pretrained weights in the model’s architecture.

The training pipeline utilizes the training split, with model performance validated on the validation split using 
metrics such as \textbf{BLEU}, \textbf{METEOR}, and \textbf{CIDEr}. To enhance variability and improve model generalization, 
we \textbf{sample one of the five} reference captions for each image during every training batch iteration, while \textbf{utilizing all five}
captions as references for metric calculation in the validation split. Hyperparameter configurations, including model 
architecture settings, are managed via a YAML configuration file. Additionally, our pipeline incorporates \textbf{grid search 
for hyperparameter optimization}, enabling systematic exploration of parameter space to identify optimal configurations. 
Training progress and performance metrics are monitored through \textbf{TensorBoard}, facilitating visualization and analysis. 
We trained each model for a total of \textbf{50 epochs} to ensure sufficient learning while avoiding overfitting.

A key efficiency enhancement in our pipeline is the \textbf{precomputation of image features}. Given that our models leverage 
pretrained image encoders, we avoid the redundant process of encoding the same images in each epoch by precomputing and 
storing the features as serialized files (pickle format). This optimization yielded a \textbf{significant reduction in training time—approximately 20x—} 
by mitigating a major computational bottleneck.

For model evaluation, we employ the optimal hyperparameter configurations determined during the tuning phase to assess 
performance on the \textbf{held-out test set}. We also used all five captions as references for each image in metric calculation. 
The evaluation process generates a detailed output comprising both quantitative metrics and qualitative results (e.g., generated 
captions), which are systematically saved in JSON files. This approach enables a comprehensive summary and facilitates 
subsequent analysis of the model’s capabilities and limitations.

\subsection{Inference Details}
We incorporate two main approaches for inference across all models: \textbf{greedy search} and \textbf{beam search} with a \textbf{beam width of 3}.

\paragraph{Greedy Search} Greedy search is a straightforward decoding strategy where, at each time step, the model selects the word with the highest predicted probability as the next word in the sequence. This approach is computationally efficient and produces results quickly. However, greedy search can sometimes lead to suboptimal captions because it does not consider multiple alternative paths that could lead to better overall sequences.

\paragraph{Beam Search} To address the limitations of greedy search, we also employ beam search with a beam width of 3. Beam search keeps track of the top 3 most probable sequences at each time step, allowing the model to explore multiple possible paths and choose a more globally optimal caption. This method results in more diverse and often more accurate descriptions by evaluating multiple potential sequences rather than just selecting the highest probability word at each step. We use the \textbf{log softmax} of the predicted probabilities as the heuristic score for evaluating candidate sequences, ensuring that the search accounts for both likelihood and sequence length in a balanced manner. The beam width of 3 strikes a balance between computational complexity and performance improvement, offering a broader search space without making inference too slow or resource-intensive.

These two methods allow us to balance between speed and output quality, enabling flexible inference based on specific requirements of efficiency or accuracy.

\subsection{Model Details}

\subsubsection{Details on CNN-RNN}

For our implementation of a CNN-RNN architecture, we utilize a pretrained \textbf{InceptionV3} as the convolutional neural network (CNN) for feature extraction. Specifically, we extract the output features from the second-to-final layer logits of InceptionV3, which provide rich, high-level visual representations of the image.

To improve efficiency and prevent overfitting, we freeze all layers of the CNN feature extractor during training, meaning that the weights of InceptionV3 are not updated. This allows the model to benefit from the pre-trained high-level image features while reducing the computational cost and ensuring stability during training.

These logits are then passed through a linear layer, and the output of the embedding size will serve as the initial input to the recurrent neural network (RNN) component, implemented using a Long Short-Term Memory (LSTM) network. Following this, we provide a Start of Sequence <SOS> token as input to the LSTM, initiating the process of generating descriptive captions for the image, one token at a time until it generates an <EOS> token.

\subsubsection{Details on CNN-Attention}

\begin{figure}[b]
  \centering
  \includegraphics[width=0.5\textwidth]{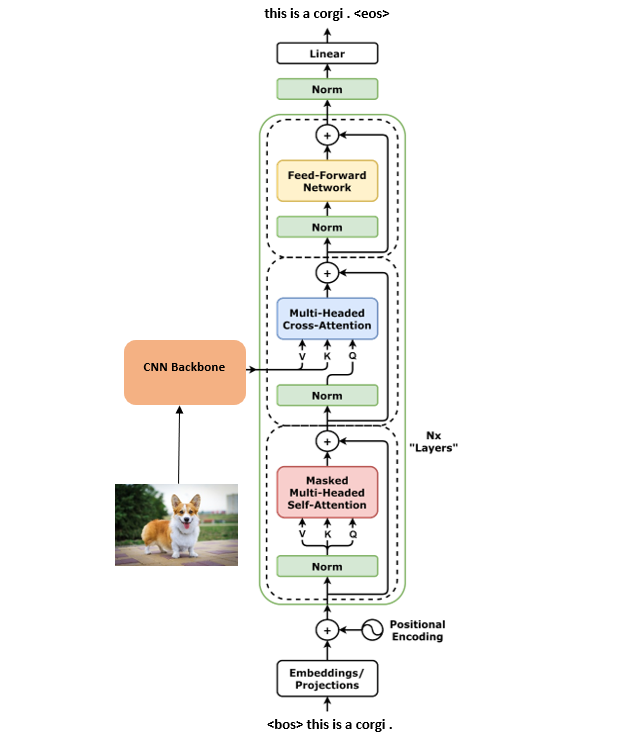}
  \caption{CNN-Attention Architecture for Image Captioning}
   \label{fig:cnn-attn}
\end{figure}

For our implementation of a CNN-Attention architecture, we utilize a similar feature extractor as our CNN-RNN architecture. Specifically, we use InceptionV3 for extracting high-level image features. However, instead of directly passing these features into an LSTM, we incorporate an attention mechanism to focus on the most relevant regions of the image for each word being generated.

We use a similar image embedding as the CNN-RNN architecture; however, this time, we use that embedding as our encoder output in a decoder-only transformer. Specifically, the image embedding is inserted as the \textbf{K (key)} and \textbf{V (value)} in the transformer architecture. This attention-based interaction between the image representation and the language model helps focus on the most relevant regions when generating each token of the caption, leading to improved descriptive power and reducing the risk of vanishing gradients.

To ensure the decoder generates captions sequentially, we apply proper masking to the output sequence so that the transformer can only attend to previously generated words and not future ones. This causal masking approach ensures that the model generates each token based solely on past information, thereby preventing it from "cheating" by looking ahead during training.

\subsubsection{Details on YOLOCNN-Attention}

For our implementation of the YOLOCNN-Attention model, we utilize the YOLO (You Only Look Once) framework for object detection, leveraging its capabilities for real-time object detection and localization. Specifically, we pass the bounding boxes concatenated with a one-hot encoded class prediction vector obtained from YOLO's detection layers. 

The output is then flattened and passed through a single layer to produce an embedding-sized vector, which is used as the key and value inputs for the Decoder block to generate captions.

\subsubsection{Details on ViT-Attention}

In this model, we utilize the Vision Transformer (ViT-16) with pretrained weights for feature extraction. ViT-16 divides the input image into patches and processes them through self-attention mechanisms, effectively capturing complex visual patterns and relationships within the image.

The final output from ViT-16 is then passed through a linear layer to produce an embedding-sized vector. This vector is used as the key and value inputs for the Transformer Decoder layer, allowing the model to attend to relevant image features and generate descriptive captions based on the visual content.

\subsubsection{Details on ViTCNN-Attention}

Finally, we proposed this \textbf{novel hybrid} combination using both ViT (Vision Transformer) and Inception CNN Architecture and stack them as encoder output. This hybrid approach leverages the complementary strengths of both architectures: the Inception CNN is adept at extracting high-level spatial features, while the Vision Transformer (ViT) excels at capturing global context through its self-attention mechanism. 

By stacking the outputs of both architectures, we create a rich and comprehensive feature representation of the image, combining both detailed spatial features and contextual relationships. The transformer decoder, utilizing these embeddings, generates descriptive captions by dynamically attending to the most relevant features at each step.

This hybrid approach is designed to enhance the model's ability to capture intricate details as well as broader contextual information, resulting in higher-quality captions. The inclusion of both local spatial and global contextual features also helps the model better handle complex scenes with multiple objects and interactions, ultimately improving the overall descriptive power and robustness of the captioning system.

\section{Experiments}
\subsection{Evaluation Results}

\begin{table}[b]
  \caption{Evaluation Results for Flickr30k and MSCOCO}
  \label{table-2}
  \centering
  \begin{tabular}{ccccccc}
    \toprule
    & \multicolumn{3}{c}{Flickr30k} & \multicolumn{3}{c}{MSCOCO} \\
    \cmidrule(r){2-4} \cmidrule(l){5-7}
    Model                   & BLEU-4 & METEOR & CIDER & BLEU-4 & METEOR & CIDER \\
    \midrule
    \textbf{CNN-RNN}        & 19.63  & 36.42  & 36.55 & 28.08  & 44.99  & 85.82 \\
    \textbf{CNN-Attn}       & 20.90  & 36.83  & 41.16 & 28.62  & 45.31  & 88.04 \\
    \textbf{YOLOCNN-Attn}   & 18.57  & 35.38  & 36.05 & 27.96  & 44.61  & 84.37 \\
    \textbf{ViT-Attn}       & 20.75  & 37.41  & 43.20 & 28.69  & 45.81  & 88.79 \\
    \midrule
    \textbf{ViTCNN-Attn}    & \textbf{21.93} & \textbf{38.26} & \textbf{44.64} & \textbf{31.24} & \textbf{46.98} & \textbf{95.30} \\
    \bottomrule
  \end{tabular}
\end{table}

\begin{figure}
  \centering
  \includegraphics[width=1\textwidth]{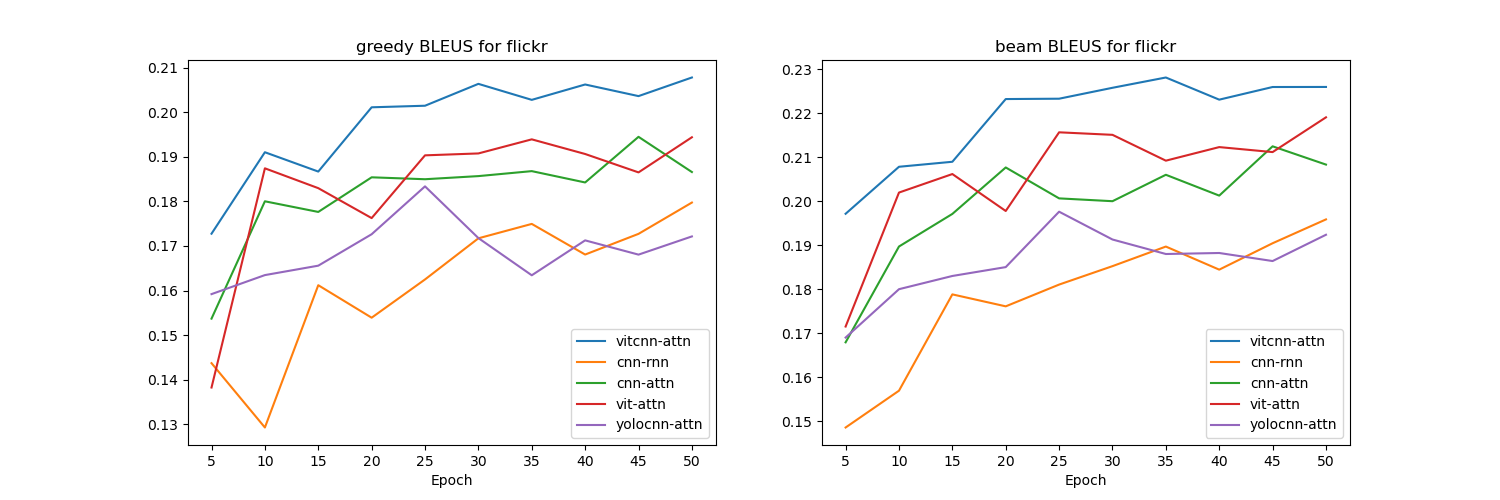}
  \caption{BLEU Plots For Flickr}
   \label{fig:bleu_flickr}
\end{figure}

\begin{figure}
  \centering
  \includegraphics[width=1\textwidth]{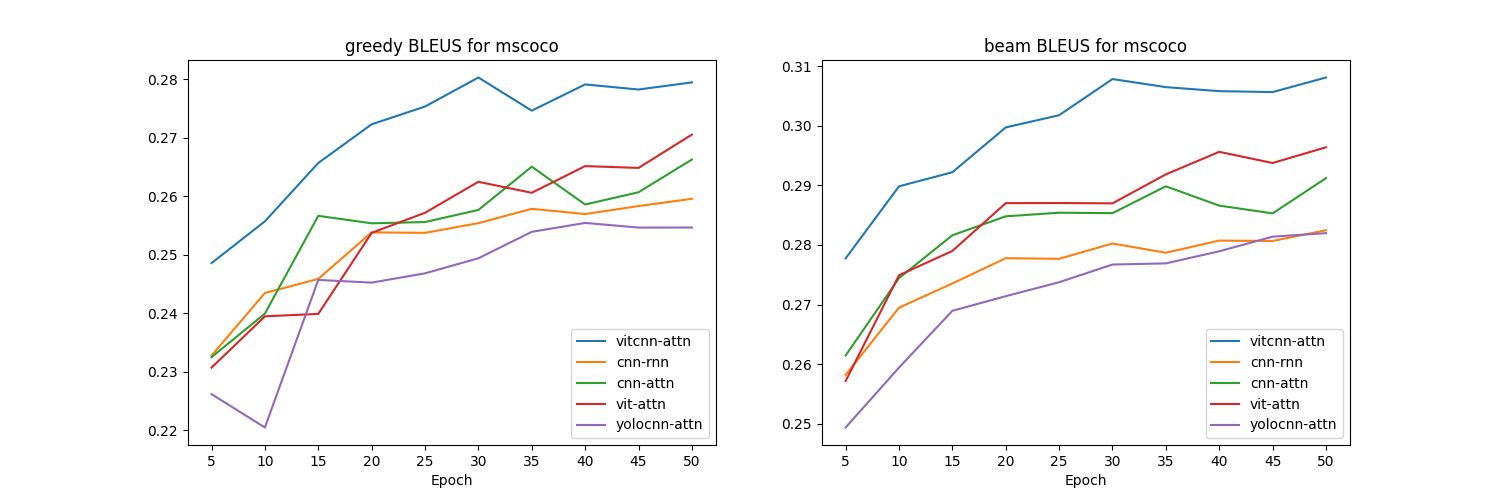}
  \caption{BLEU Plots For MSCOCO}
   \label{fig:bleu_mscoco}
\end{figure}

In this section, we present the evaluation results of our models on the Flickr30k and MSCOCO test datasets. Table \ref{table-2} summarizes the performance metrics, including BLEU-4, METEOR, and CIDEr scores, for the different model architectures used in our experiments. Additionally, Figures \ref{fig:bleu_flickr} and \ref{fig:bleu_mscoco} illustrate the progression of BLEU scores on the validation split over different epochs, providing insights into the models' training dynamics and performance trends.

We experimented with various model architectures, including CNN-RNN, CNN-Attn, ViT-Attn, ViTCNN-Attn, YOLO-Attn, and YOLOCNN-Attn. Additionally, we explored using variational autoencoders (VAE) by integrating image encoders with models such as VAE-YOLO and VAE-CNN with attention mechanisms. However, these VAE-based models were excluded from the final results due to their subpar performance compared to the other architectures tested.

The \textbf{ViTCNN-Attn} model outperformed all others, achieving the highest scores on both datasets. Its superior performance over \textbf{ViT-Attn} and \textbf{CNN-Attn} highlights the effectiveness of combining ViT's global context understanding with CNN's spatial feature extraction, leveraging both local and global information for better captions.

In contrast, \textbf{CNN-RNN} struggled more with long-range interactions, and \textbf{YOLOCNN-Attn} models fell short in capturing more complex spatial relations. Overall, the \textbf{ViTCNN-Attn} model demonstrates a robust approach to image captioning, achieving great results across metrics.

\subsection{Hyper-parameter Tuning}

\begin{table}
  \caption{Best Hyperparameters For Each Model}
  \label{table-3} 
  \centering
  \begin{tabular}{ccccccccc}
    \toprule
    & \multicolumn{4}{c}{Flickr30k} & \multicolumn{4}{c}{MSCOCO} \\
    \cmidrule(r){2-5} \cmidrule(l){6-9}
    Model                   & bs & lr & es & nl & bs & lr & es & nl  \\
    \midrule
    \textbf{CNN-RNN}        & 128 & 0.0005 & 256 & 1  & 128 & 0.0005      & 256 & 1 \\
    \textbf{CNN-Attn}       & 64  & 0.001  & 512 & 1  & 64  & 0.0005      & 512 & 1 \\
    \textbf{YOLOCNN-Attn}   & 64  & 0.001  & 256 & 2  & 128 & 0.001       & 512 & 1 \\
    \textbf{ViT-Attn}       & 64  & 0.0005 & 512 & 2  & 64  & 0.001       & 512 & 1 \\
    \textbf{ViTCNN-Attn}    & 64  & 0.0005 & 256 & 1  & 64  & 0.001       & 256 & 1 \\
    \bottomrule
  \end{tabular}
\end{table}

In our experiments, we performed extensive hyper-parameter tuning to optimize the model's performance. Specifically, we adjusted several key parameters, including the \textbf{batch size}, \textbf{learning rate}, \textbf{embedding size}, and the \textbf{number of layers} for attention-based models. 

\begin{enumerate}
    \item \textbf{Batch Size (64, 128)}: We experimented with different batch sizes to balance convergence speed and stability, finding the optimal trade-off between faster training and effective generalization.
    \item \textbf{Learning Rate (1e-3, 5e-4)}: The learning rate was fine-tuned to ensure efficient convergence, preventing both underfitting (due to low rates) and instability (due to excessively high rates). 
    \item \textbf{Embedding Size (256, 512)}: The size of the word embedding was varied to evaluate its impact on capturing the semantic nuances of the captions while ensuring computational feasibility.
    \item \textbf{Number of Layers for Attention-based Models (1,2,4)}: For models using \textbf{attention mechanisms}, the number of encoder-decoder layers was carefully tuned to enhance the model’s ability to focus on relevant regions of the input image, improving the quality of the generated descriptions.
\end{enumerate}

The tuning process aimed to achieve the best possible balance between training efficiency and model performance, allowing us to identify hyper-parameter settings that yielded optimal results across our evaluation metrics.

Table \ref{table-3} shows the parameters with the highest \textbf{BLEU Score} using \textbf{Beam Search} with a width of 3.

\subsection{Qualitative Analysis}

In this section, we conduct a qualitative analysis by examining the captions generated by each model after optimal hyperparameter tuning. We present a selection of images where the models produce both similar and distinct captions, using unseen test data from the Flickr30k and MS-COCO datasets.

\begin{figure}[h!]
    \centering
    \begin{minipage}{0.45\textwidth}
        \centering
        \includegraphics[width=1\linewidth]{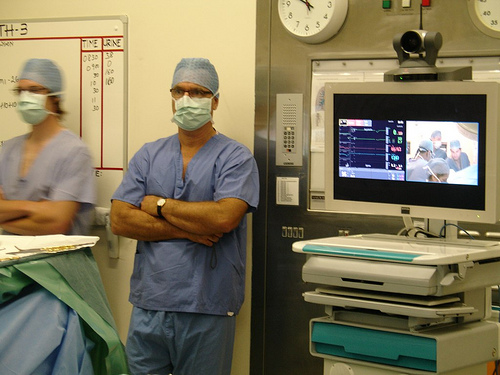} 
    \end{minipage}%
    \hspace{0.001\textwidth}%
    \begin{minipage}{0.45\textwidth}
        \begin{itemize}
            \item \textbf{CNN-RNN:} "A man in a white shirt is standing in a kitchen."
            \item \textbf{CNN-Attn:} "A man in a blue shirt is standing in front of a large television."
            \item \textbf{YOLOCNN-Attn:} "A man in a blue shirt is standing in front of an ATM machine."
            \item \textbf{ViT-Attn:} "Two medical professionals are examining a medical procedure."
            \item \textbf{ViTCNN-Attn:} "A man in a blue shirt is standing in front of a computer."
            
        \end{itemize}
    \end{minipage}
    \caption{Image with different generated captions from different models.}
    \label{fig:model_captions_2}
\end{figure}

In figure \ref{fig:model_captions_2}, caption generation varied significantly across models. Most models successfully identified the man present in the image; however, only the ViT-Attn model accurately recognized two men instead of one. The interpretations of the background also varied considerably. The ViT-Attn model produced the most accurate caption overall, followed by ViTCNN-Attn and CNN-Attn, which described the man as being next to a computer or television. In contrast, the CNN-RNN and YOLOCNN-Attn models misidentified objects surrounding the man, leading to less accurate descriptions.

\begin{figure}[h!]
    \centering
    \begin{minipage}{0.45\textwidth}
        \centering
        \includegraphics[width=0.8\linewidth]{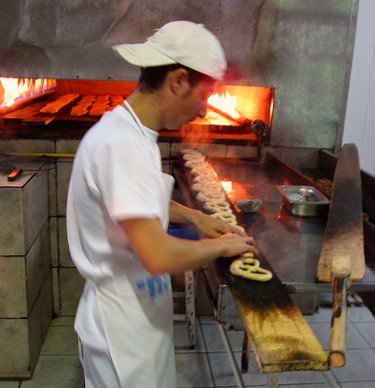} 
    \end{minipage}%
    \hspace{0.001\textwidth}%
    \begin{minipage}{0.45\textwidth}
        \begin{itemize}
            \item \textbf{CNN-RNN:} "a man in a white shirt is standing in front of a grill"
            \item \textbf{CNN-Attn:} "a man in a white shirt and tie is preparing food"
            \item \textbf{YOLOCNN-Attn:} "a man in a white shirt is preparing food"
            \item \textbf{ViT-Attn:} "a chef is preparing food in a kitchen"
            \item \textbf{ViTCNN-Attn:} "a man in a white apron is preparing food"
           
        \end{itemize}
    \end{minipage}
    \caption{Image with similar generated captions from different models.}
    \label{fig:model_captions_1}
\end{figure}

\section{Limitations and Possible Improvements}

Our approach has several limitations, mainly related to computational resources and training time. The experiments were carried out using V100 and RTX 4060 GPUs, which constrained both the size of the model and the extent of hyperparameter tuning we could perform. These limitations impacted our ability to explore more complex architectures and fully optimize performance.

Possible improvements include increasing the model capacity by leveraging more powerful hardware, allowing for the training of larger models and deeper hyperparameter exploration. Additionally, experimenting with different combinations of existing models, as well as incorporating state-of-the-art methods such as GPT models and BLIP (Bootstrapped Language-Image Pretraining), could significantly enhance the model's capabilities and overall performance.

\section{Conclusion}
In this project, we developed an automated image captioning system that evolved through progressively advanced models, starting with CNN-RNN and culminating in a hybrid Vision Transformer-CNN with Attention (ViTCNN-Attn). The initial CNN-RNN approach provided a foundational understanding of how visual features can be extracted and used to generate descriptive captions. By incorporating attention mechanisms into CNN-Attn models, we enhanced the system's ability to focus on important parts of the embeddings, leading to more contextually relevant and detailed captions. The integration of Vision Transformers (ViT) further improved the model's ability to capture complex visual relationships, offering more accurate and expressive captions. We have shown that these additions have led to an increase in performance.

Finally, we proposed a \textbf{ new hybrid ViTCNN-Attn} model, combining the strengths of both CNNs and Vision Transformers, and demonstrated a significant improvement in generating captions that captured both local features and global context effectively. The performance of these models was evaluated using metrics such as BLEU, METEOR, and CIDEr, which confirmed the improvements gained by the introduction of advanced architectures and attention mechanisms.

\newpage
\bibliographystyle{unsrt}
\bibliography{references}


\newpage
\appendix

\section{Appendix}

\subsection{Flickr Dataset: Training Loss and Validation Metrics}

\begin{figure}[ht]
    \centering
    \includegraphics[width=\textwidth]{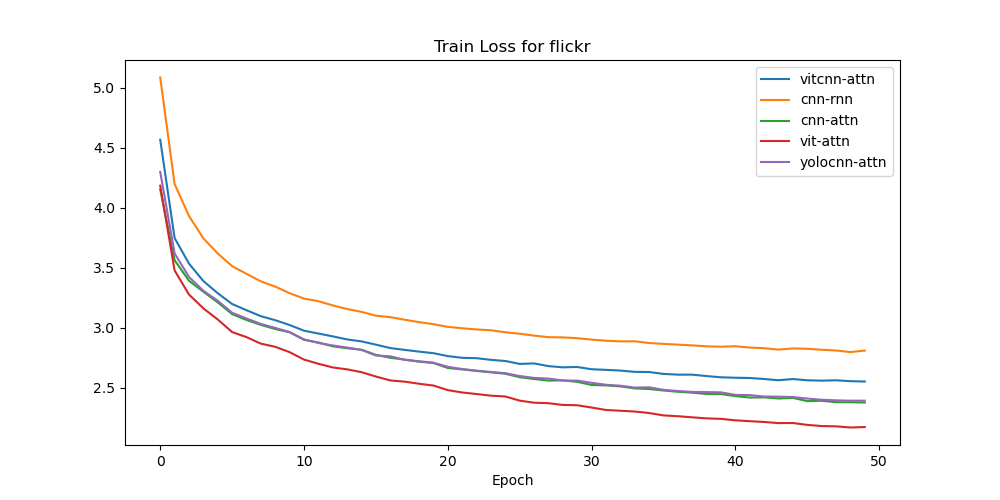}
    \caption{Training Loss Plots for Flickr Dataset}
    \label{fig:flickr_loss_plot}
\end{figure}

\begin{figure}[ht]
    \centering
    \includegraphics[width=\textwidth]{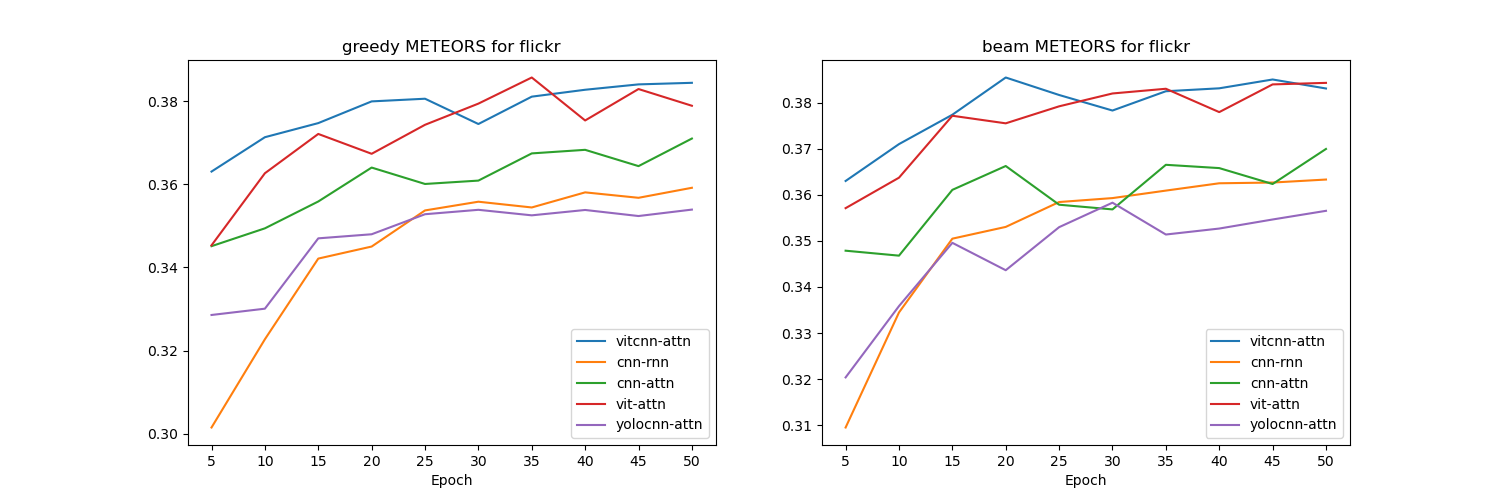}
    \caption{METEOR Plots for Flickr Dataset}
    \label{fig:flickr_meteor_plot}
\end{figure}

\begin{figure}[ht]
    \centering
    \includegraphics[width=\textwidth]{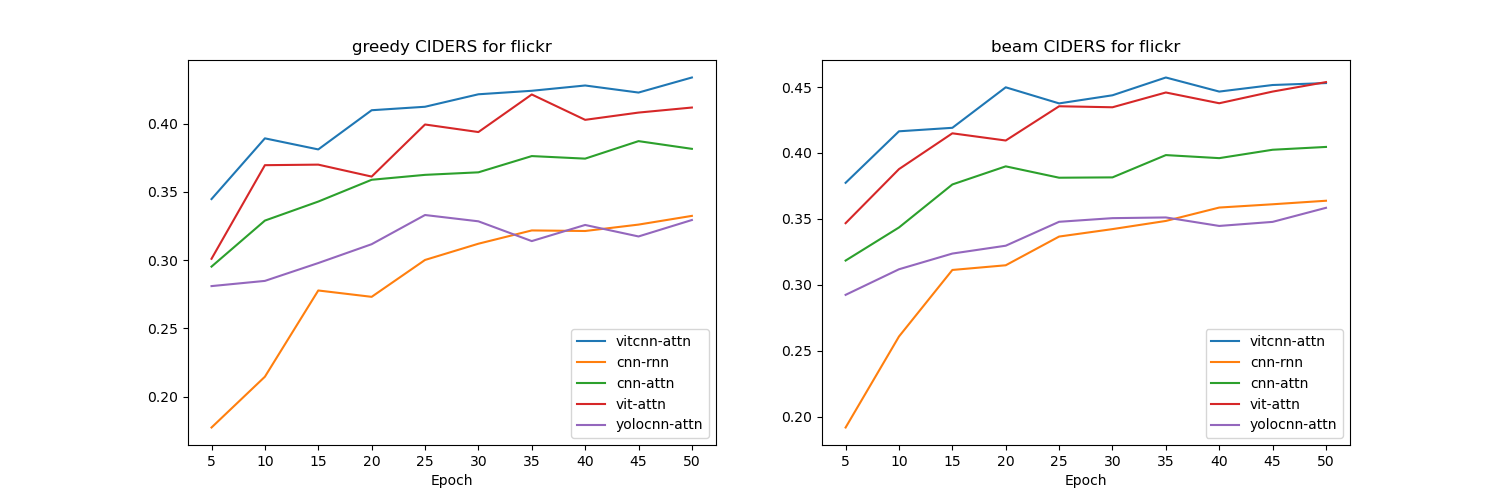}
    \caption{CIDEr Plots for Flickr Dataset}
    \label{fig:flickr_cider_plot}
\end{figure}

\newpage
\subsection{MSCOCO Dataset: Training Loss and Validation Metrics}

\begin{figure}[ht]
    \centering
    \includegraphics[width=\textwidth]{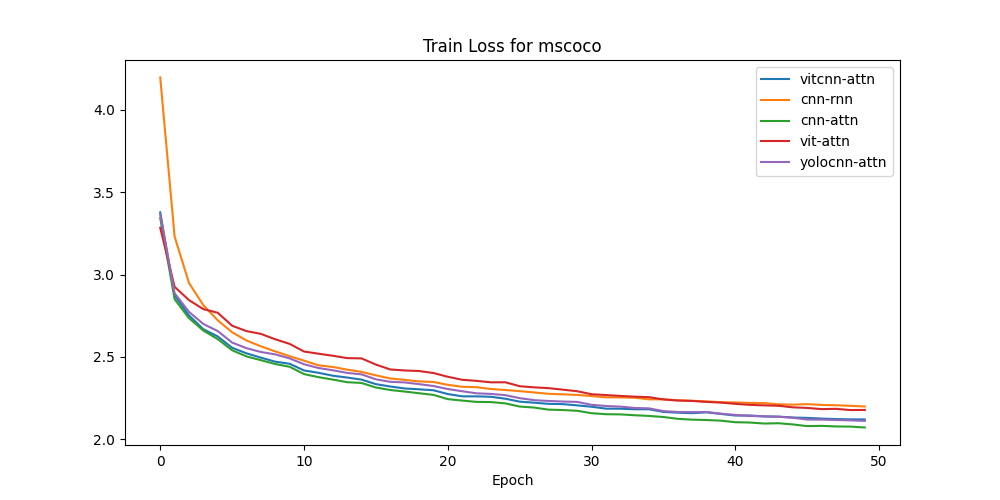}
    \caption{Training Loss Plots for MSCOCO Dataset}
    \label{fig:mscoco_loss_plot}
\end{figure}

\begin{figure}[ht]
    \centering
    \includegraphics[width=\textwidth]{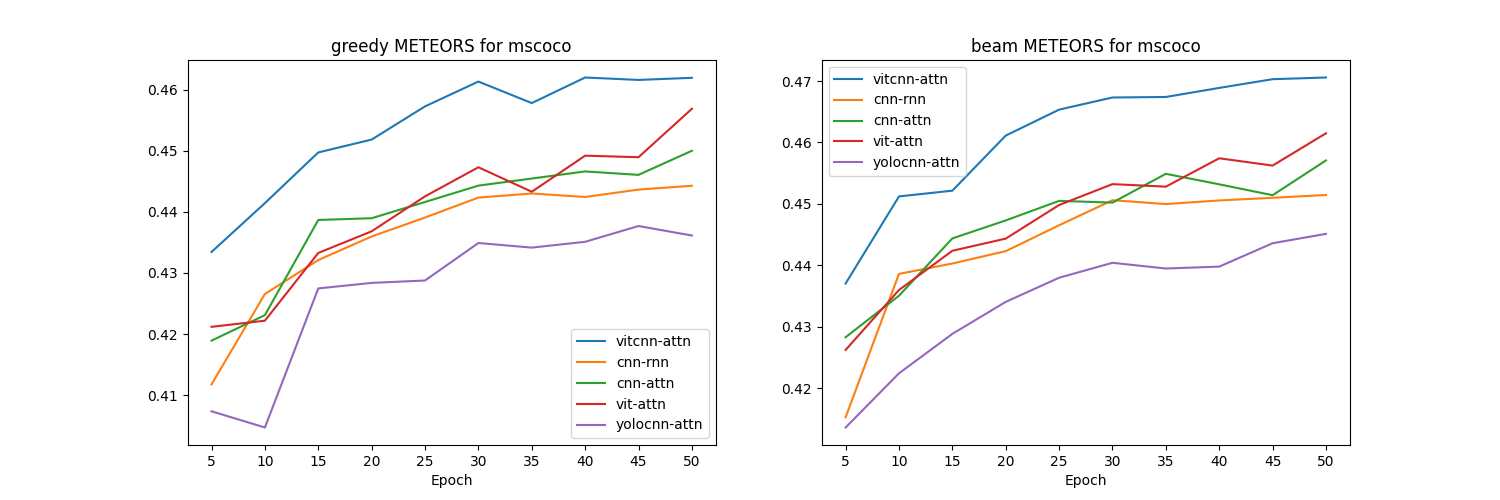}
    \caption{METEOR Plots for MSCOCO Dataset}
    \label{fig:mscoco_meteor_plot}
\end{figure}

\begin{figure}[ht]
    \centering
    \includegraphics[width=\textwidth]{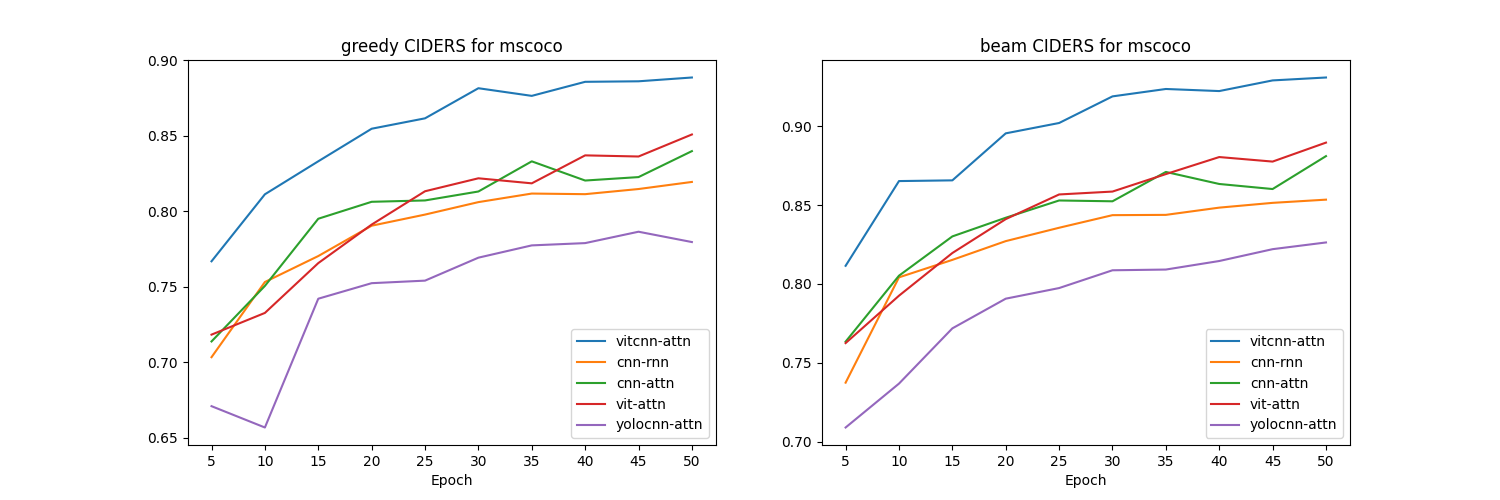}
    \caption{CIDEr Plots for MSCOCO Dataset}
    \label{fig:mscoco_cider_plot}
\end{figure}


\end{document}